\documentclass[border=3mm]{article}
\usepackage{ijcai25}

\usepackage[table,xcdraw]{xcolor} 
\usepackage{times}
\usepackage{soul}
\usepackage{url}
\usepackage[hidelinks]{hyperref}
\usepackage[utf8]{inputenc}
\usepackage[small]{caption}
\usepackage{graphicx}
\usepackage{amsmath}
\usepackage{amsthm}
\usepackage{amssymb}
\usepackage{booktabs}
\usepackage{algorithm}
\usepackage{algorithmic}
\usepackage{arydshln}
\usepackage[switch]{lineno}

\usepackage{tikz}
\usepackage{multicol}
\usepackage{adjustbox}
\usepackage{lipsum}
\usepackage{threeparttable}

\usepackage{makecell}
\usepackage{etoolbox}

\newcolumntype{C}[1]{>{\centering\arraybackslash}m{#1}}

\newcommand{\icon}[2][0.2in]{%
  \raisebox{-0.1\height}{\includegraphics[width=#1]{#2}}%
}

\newcommand{\pickIconNew}[1][0.12in]{%
  \icon[#1]{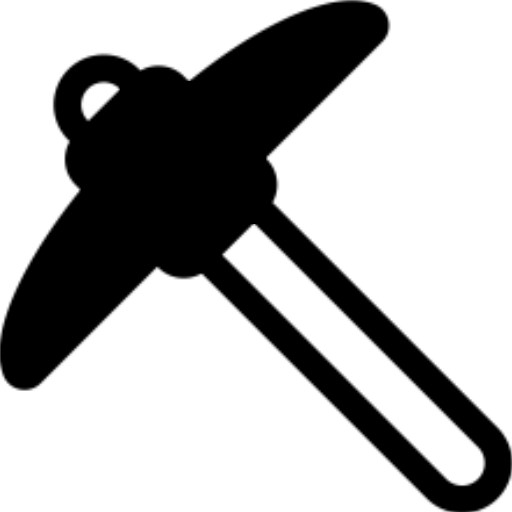}%
}

\newcommand{\anchorIconNew}[1][0.12in]{%
  \icon[#1]{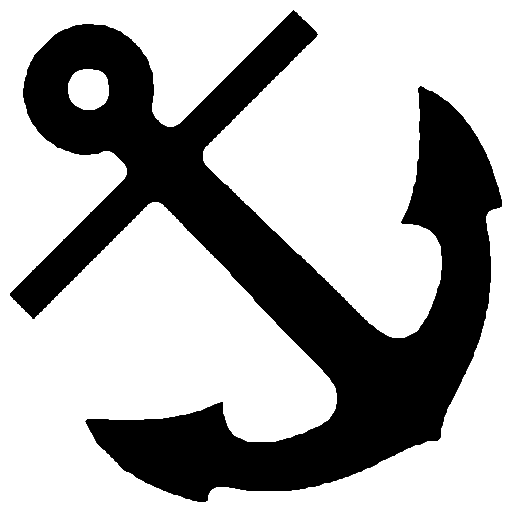}%
}

\newcommand{\potionIcon}[1][0.125in]{%
  \icon[#1]{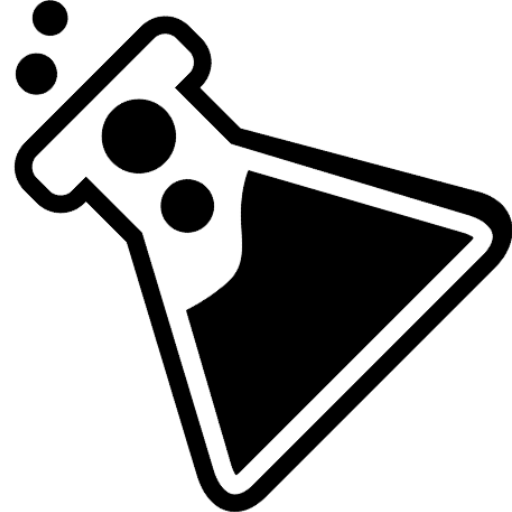}%
}

\usepackage{array, colortbl}
\newcolumntype{C}[1]{>{\centering\arraybackslash}p{#1}}
\newcolumntype{L}[1]{>{\raggedright\arraybackslash}p{#1}}
\newcolumntype{R}[1]{>{\raggedleft\arraybackslash}p{#1}}

\usepackage{tikz}
\usepackage{wheelchart}
\usepackage {siunitx} 
\usepackage{comment}

\usetikzlibrary{decorations.markings}
\usetikzlibrary{decorations.text}
\usetikzlibrary{patterns}

\definecolor{Apricot}{RGB}{251,206,177}
\definecolor{LimeGreen}{RGB}{50,205,50}
\definecolor{Melon}{RGB}{253,188,180}
\definecolor{OliveGreen}{RGB}{107,142,35}
\definecolor{Peach}{RGB}{255,229,180}
\definecolor{Plum}{RGB}{221,160,221}
\definecolor{WildStrawberry}{RGB}{255,67,164}
\definecolor{DarkOrchid}{RGB}{153,50,204}

\definecolor{oldbrick}{HTML}{966c6c}
\definecolor{fwhite}{HTML}{ffffff}
\definecolor{japreddo}{HTML}{e83b3b}
\definecolor{freshbrick}{HTML}{cd683d}
\definecolor{darkyellow}{HTML}{f9c22b}
\definecolor{illsapgreen}{HTML}{cddf6c}
\definecolor{pinegreen}{HTML}{0b5e65}
\definecolor{deepurple}{HTML}{6b3e75}
\definecolor{hazepurple}{HTML}{a24b6f}
\definecolor{floydpurple}{HTML}{905ea9}
\definecolor{avgpink}{HTML}{ed8099}
\definecolor{palestpink}{HTML}{fdcbb0}
\definecolor{mandarin}{HTML}{f79617}

\definecolor{smoothiepurple}{HTML}{eaaded}
\definecolor{tittypink}{HTML}{f04f78}

\definecolor{mclovingreen}{HTML}{0eaf9b}

\definecolor{lightgrey}{HTML}{bbbbbb}
\definecolor{darkergrey}{HTML}{777777}
\definecolor{darkestgrey}{HTML}{555555}
\usetikzlibrary{patterns}

\gdef\venueWheel{%
50 (20)/pinegreen!95/AAAI,
31 (11)/illsapgreen!95/IJCAI,
28 (17)/darkyellow!95/NeurIPS,
17 (10)/mandarin!95/ICLR,
17 (14)/freshbrick!95/ICML,
9 (6)/oldbrick!95/EMNLP,
7 (4)/japreddo!95/CVPR,
7 (5)/tittypink!95/ACL,
2 (0)/avgpink!95/ICCV,
2 (2)/smoothiepurple!95/JMLR,
1 (1)/hazepurple!95/NAACL,
1 (1)/deepurple!95/TACL
}

\gdef\inclusionCrit{92/mclovingreen!95/NeSy/mclovingreen!95,
42/lightgrey!35/Black-Box/darkergrey!90,
20/darkergrey!35/Logic/darkergrey!95, 
18/darkestgrey!35/Short/darkestgrey!95
}

\newcommand{\blackBox}{black-box~}

\def\acceptedlegendColor{mclovingreen!95}
\def\discardedlegendColor{lightgrey!65}

\title{\vspace{-1.2cm}Neuro-Symbolic Artificial Intelligence:\\A Task-Directed Survey in the Black-Box Models Era\vspace{-0.3cm}}

\author{%
Giovanni Pio Delvecchio$~^1$\and 
Lorenzo Molfetta$~^1$\and 
Gianluca Moro$~^1$\\
\affiliations Department of Computer Science and Engineering, University of Bologna, Cesena Campus\\
Via dell'Università 50, I-47522 Cesena, Italy\\
\emails \{g.delvecchio, lorenzo.molfetta, gianluca.moro\}@unibo.it
}

\begin{document}

\maketitle

\def\thefootnote{1}\footnotetext{Equal contribution (co-first authors).}\def\thefootnote{\arabic{footnote}}
\def\thefootnote{*}\footnotetext{The definitive, copyrighted, peer-reviewed, and edited version of this Article is published in \textit{IJCAI 2025}, pp. 10418-10426, 2025. DOI: \url{https://doi.org/10.24963/ijcai.2025/1157}.}\def\thefootnote{\arabic{footnote}}

\begin{abstract}
The integration of symbolic computing with neural networks has intrigued researchers since the first theorizations of Artificial intelligence (AI). 
The ability of Neuro-Symbolic (NeSy) methods to infer or exploit behavioral schema has been widely considered as one of the possible proxies for human-level intelligence.
However, the limited semantic generalizability and the challenges in declining complex domains with pre-defined patterns and rules hinder their practical implementation in real-world scenarios.
The unprecedented results achieved by connectionist systems since the last AI breakthrough in 2017 have raised questions about the competitiveness of NeSy solutions, with particular emphasis on the Natural Language Processing and Computer Vision fields.
This survey examines task-specific advancements in the NeSy domain to explore how incorporating symbolic systems can enhance explainability and reasoning capabilities.
Our findings are meant to serve as a resource for researchers exploring explainable NeSy methodologies for real-life tasks and applications.
Reproducibility details and in-depth comments on each surveyed research work are made available at \url{https://github.com/disi-unibo-nlp/task-oriented-neuro-symbolic.git}.
\end{abstract}

\section{Introduction}

``\textit{Stacked neural layers is all we need}''. This statement summarizes most of the current research efforts in the Artificial Intelligence (AI) field, especially in Natural Language Processing (NLP) and Computer Vision (CV).
The advancements and achievements of the latest neural models are marvelous, but they conceal fundamental drawbacks regarding data efficiency and explainability.
The need to resort to Neuro-Symbolic (NeSy) components naturally arises from the necessity for trustworthy and efficient solutions.
\begin{figure}[!ht]
    \centering
    \resizebox{6.7cm}{!}{
    \begin{tikzpicture}
      \def\WCtest#1#2{%
        \pgfmathparse{true?"#1":"#2"}%
        \pgfmathresult%
      }
      
      \wheelchart[
        slices style={draw=black, fill=\WCvarB},
        data={},
        legend columns=1,
        legend row={\tikz\fill[\WCvarB] (0,0) rectangle (0.3,0.3); & \WCvarC & \WCvarA},
        legend={\node[anchor=north] at (7,3.2) {\resizebox{5cm}{!}{%
            \begin{tabular}{*{4}{l@{ }lr}}\WClegend\end{tabular}}};},
        radius={3.5}{4.5}
      ]{\venueWheel}
      
    \wheelchart[
      slices style={
        draw,
        fill=none,                     
        preaction={fill, fill=\WCvarB}, 
        postaction={pattern=\WClistpattern, pattern color=black!40}
      },
      arc=\WCvarB,
      arc around line=1,
      arc around text,
      arc data=\WCvarC,         
      wheel data=\WCvarA,       
      wheel data style={font=\Large},
      wheel data pos=0.5,       
      arc data dir={\WCmidangle<180?1:-1},
      arc data expand=f,
      arc data pos=0.0,
      arc data style={text color=\WCvarD},
      arc first half=-,
      arc pos=0.0,
      arc second half=-,
      data={},                 
      radius={2}{3},
      WClistpattern={
        none, north east lines, north east lines, north east lines
      }
    ]{\inclusionCrit}

    \node[anchor=north] at (0,-5.0) {%
        \begin{tabular}{@{} >{\centering\arraybackslash}m{1cm}  l  >{\centering\arraybackslash}m{1cm}  l @{}}
          \tikz[baseline]{%
             \fill[\acceptedlegendColor] (0,0) rectangle (0.5,0.5);
          }
          & {\huge Considered}
          &
          \tikz[baseline]{%
             \fill[\discardedlegendColor] (0,0) rectangle (0.5,0.5);
             \begin{scope}
               \clip (0,0) rectangle (0.5,0.5);
               \fill[pattern=north east lines, pattern color=black!50, line width=0pt] (0,0) rectangle (0.5,0.5);
             \end{scope}
          }
          & {\huge Discarded}
        \end{tabular}
      };
      
    \end{tikzpicture}
    }
    \caption{Distribution of NeSy peer-reviewed papers over the period 2017-2024.
    The innermost ring delineates the inclusion criteria. Greyscale slices denote studies excluded for relying on black‑box methods, exclusively logical approaches, or brevity, while the colored slice comprises the surveyed ones.
    The outermost ring represents the number of research works from each selected venue, with exact counts reported in the legend.
    The number of papers considered for each track is enclosed in parentheses.
    } 
    \label{fig:placeholder_circles}
\end{figure}
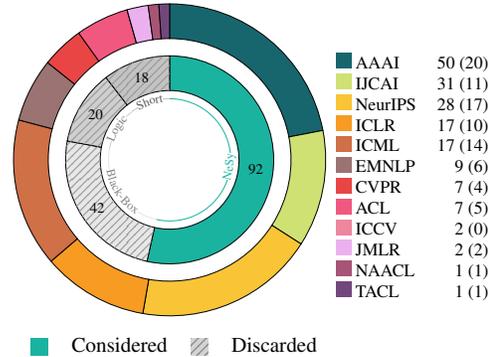
%
The data type of ``thoughts'' in the connectionist approaches---also known as tensors---and the hidden unsupervised manipulations of such information constitute a physical barrier to hierarchical and abstract planning, whose achievement cannot be reached via mere input-output relationships~\cite{DBLP:journals/corr/abs-1801-00631}. 
If we envision a not-too-distant future where data availability and quality become critical concerns, mainly due to the potential bias introduced by synthetic information, we urge finding and exploring research pathways with compositional abilities weakly related to the training set's dimension~\cite{DBLP:conf/ijcai/GiunchigliaSL22}.
We uphold that neuro-symbolic methods can provide an alternative to break the curse-of-dimensionality modern methodologies suffer from.
While effective, sparsity enforcement regularization mechanisms~\cite{DBLP:journals/jmlr/Bach17} require strong assumptions about the nature of the target distribution, which is unrealistic for the complex domains NLP and CV models usually deal with~\cite{DBLP:journals/corr/abs-2104-13478}. 
We argue that symbolic components can be the regularization for forcing more complex behavior into neural models.
By combining data-driven insights with explainable, logic-based representations, we can achieve higher generalization capabilities~\cite{DBLP:series/faia/BesoldGBBDHKLLPPPZ21}. 
Other data-driven paradigms have also been considered~\cite{DBLP:conf/adc/LodiMS10,DBLP:conf/ic3k/DomeniconiMPP15}.

As the keywords ``reasoning'' and ``explainability'' are gaining more attention in the community, we highlight a key notational misunderstanding between rule-guided and post-inference forensic strategies.
While both methods aim to provide a rationale behind predictions, they fundamentally differ in their approach. 
Rule-guided methods constrain the output to meet specific criteria, while post-inference approaches retrospectively reconstruct plausible interpretable reasoning pathways that led to the results.
We further claim that these terms should not be used to advocate human-like faculties but as proxies for generalization behavior within a hierarchical backbone architecture governed by predefined rules or component-interaction schemas.

\paragraph{Our Contribution}
The lack of a clear definition of NeSy in the AI field, the inconsistency in evaluation benchmarks, and the multitude of uncoordinated study directions with no common comparison grounds primarily drive this survey. 
In this work, we conduct a task-directed literature analysis focusing on how NeSy methods apply and scale in diverse application contexts.
We assess the methodological soundness of such approaches and compare them to \blackBox systems in real-world scenarios to identify the current limitations of these strategies.
Departing from recent surveys that attempt to classify research works along taxonomical axes or network architectural variations, we propose a new methodological approach that disentangles these conceptual and practical inconsistencies by providing a comprehensive and critical overview of NeSy systems.
We aim to deepen the understanding of these hybrid systems' advantages and inherent limitations, laying the groundwork for future research and bridging the gap between explainability and performance in AI.

\section{Taxonomy}
This section discusses the structure and systematic reproducible methodologies followed to define our task-directed taxonomy for NeSy methods. 
We further examine datasets, evaluation benchmarks, and their limitations in delineating fair comparisons across different approaches.
As previously discussed, we acknowledge the nuanced usages and inconsistency in the declination of the term NeSy in NLP and CV literature, where it is often misapplied to methods claiming reasoning and interpretability without structured symbolic frameworks. 
In this paper, NeSy exclusively refers to approaches integrating neural networks with symbolic components such as solvers, logical rules, or state-action schemas.

\paragraph{Inclusion Criteria}
This survey aims to investigate NeSy systems from the last AI revolution in 2017. 
To identify relevant research systematically, we used the DBLP SPARQL endpoint to collect research papers published between 2017 and 2024 from general-purpose leading venues in those fields where such technological leap has taken hold more --- Natural Language Processing (NLP) and Computer Vision (CV).
A customized query was executed to retrieve publications related to NeSy methods using the keywords: ``neuro-symbolic'', ``nesy'', ``rule-based'', ``probabilistic-logic'', ``probabilistic-reasoning'',  ``logic-based'', ``soft-logic'', ``fuzzy-logic'',  ``concept-learning'', ``inductive-logic programming''.
Figure \ref{fig:placeholder_circles} illustrates the number and distribution of the research works meeting our requirements. 
We meticulously analyzed the resulting 172 papers and categorized them to identify works that deviated from the designated NeSy formulation, excluding non-interpretable methods, purely logical approaches, and short papers.
This collection was thoroughly studied to derive a pool of research works focusing on NeSy approaches that integrate symbolic reasoning with neural networks.
We further explored relevant related works to gain a precise picture of the NeSy research landscape. 

\begin{figure*}
    \centering
    \includegraphics[width=\textwidth]{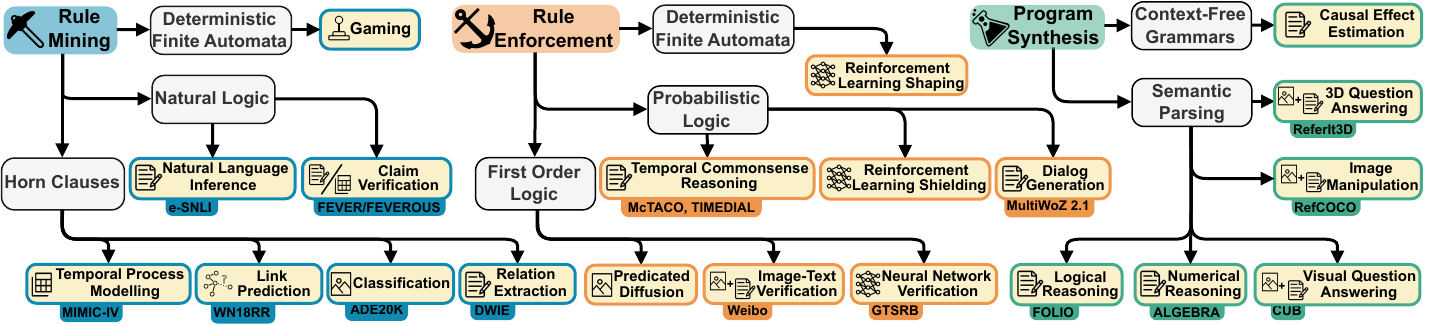}
    \caption{Task-Directed NeSy Taxonomy. 
    We organize the most relevant task families by classifying each according to applicable NeSy techniques and grouping them into three macro-categories: (1) Rule Mining, (2) Rule Enforcement, and (3) Program Synthesis. 
    We also label the most commonly used datasets to highlight their role in real-world benchmarking and model evaluation.
    Tasks are organized from bottom to top and from left to right within each category, mirroring the survey's conceptual progression.
    }
    \label{fig:placeholder_tax}
\end{figure*}

\paragraph{Datasets and Benchmarks}
We analyzed real-world datasets and benchmarks to ensure an unbiased comparison across methods while focusing on practical applications. 
This study highlighted reproducibility challenges in tasks involving sampling operations, such as negative-example mining.
We found substantial inconsistencies on the WN18RR\footnote{https://github.com/TimDettmers/ConvE} benchmark for different research works using the same \blackBox model baselines.
We further spotlight an evaluation trend in the Nesy literature where synthetic datasets and toy tasks are specifically tailored for the features of the newly proposed method.
For this reason, we consider them unsuitable for drawing conclusions and comparisons. 
In Figure \ref{fig:placeholder_tax}, we label most of the identified NeSy tasks with their respective evaluation benchmark if they respect our fairness criteria.
Missing values appear in image generation, where different instruction prompts significantly impact results; in Reinforcement Learning (RL), where each method is evaluated on custom tasks or games; and in causal effect estimation, due to outdated benchmarks with limited entries.

\begin{figure*}
    \centering
    \includegraphics[width=\textwidth]{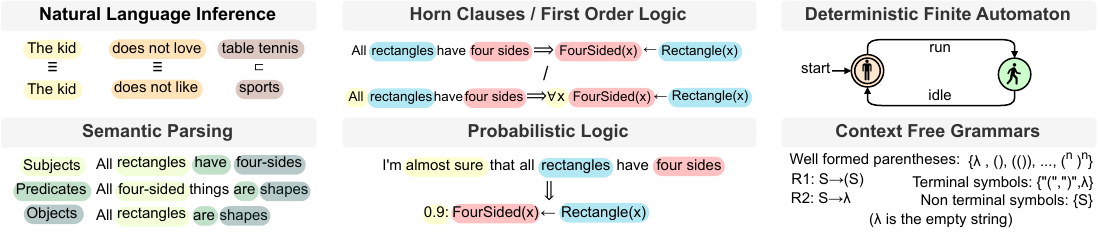}
    \caption{Overview of the intermediate formal languages employed by each taxonomy method.}
    \label{fig:formal_languages}
\end{figure*}

\paragraph{A Task-Oriented Formulation}
We organize our analysis by identifying three distinct methodological frameworks that traverse the neural-symbolic research landscape: (1) Rule Mining, (2) Rule Enforcement, and (3) Program Synthesis. 
Within each framework, we categorize techniques based on their theoretical grounding and the tasks they address, focusing on how constraints, symbolic structures, and interpretable programmatic constructs are respectively extracted, integrated, or synthesized.
Our taxonomy in Figure \ref{fig:placeholder_tax} is designed to be read in two complementary ways. 
A top-down perspective highlights how these frameworks interconnect, emphasizing shared logical formalisms and learning paradigms. 
A bottom-up approach begins with a specific task, moving upward to identify the most relevant methodological families and techniques for that objective.
This dual perspective preserves the field’s theoretical coherence while promoting a task-directed formulation of NeSy solutions. 
Enabling bidirectional exploration between related methodologies allows researchers to identify analogous strategies to refine or expand existing approaches based on their objectives.
In Figure \ref{fig:formal_languages} we show examples of the formal languages employed by each theoretical framework in the taxonomy.\\

\indent
In the following sections, we analyze each methodology in depth, illustrating its core principles and highlighting opportunities for advancements in neural-symbolic integration.

\section{Rule-Mining techniques \pickIconNew[0.15in]}
Neuro-Symbolic techniques employ rule-mining as a core methodology for model construction, focusing on extracting interpretable rules from diverse input data. 
The computational complexity of rule extraction directly correlates with the chosen formal specification language. 
Horn clause approaches exhibit the highest complexity, requiring matrix representations for ground atom truth values. 
Natural Logic frameworks show moderate complexity, combining linguistic preprocessing with question-answering for entity and operator extraction. 
Deterministic Finite Automata (DFAs) present the lowest complexity, leveraging operational traces with SAT-solving for rule learning.

\subsection{Horn Clauses}
Horn Clauses (HCs), a fundamental component of logic programming, provide a structured framework for rule-based reasoning. 
Defined as disjunctions of literals with at most one positive literal as the clause head, HCs enable logical inference across various domains.\\
\indent
Temporal point process modeling relies on HC-based reasoning to capture event dependencies, which is particularly crucial in fields such as medicine and autonomous driving, where sudden changes in conditions can have severe consequences. 
\cite{DBLP:conf/icml/YangYLF024} introduced a NeSy rule induction method that leverages a sequential covering algorithm and a custom attention mechanism to extract HCs. 
While demonstrating scalability, trustworthiness, and strong performance over existing models, its ability to capture increasingly complex rules remains an open research question.\\
\indent
The Inductive Logic Programming (ILP) paradigm automates the discovery of HC-based rules, supporting generalized reasoning and inference over structured data.
In the context of extracting rules from structured data like tables and graphs, ILP seeks to identify missing connections between entities by uncovering patterns in existing relationships.
DFORL~\cite{DBLP:journals/ai/GaoICW24}, a recent lightweight method for efficient rule extraction, introduces a depth-limited breadth-first search for neighborhood extraction. 
This propositionalization technique converts relational facts into vector representations suitable for neural network-based learning and applies syntactic constraints to reduce the rule search space. 
By integrating auxiliary matrices and curriculum learning, DFORL uncovers hidden predicates and enhances the efficiency and progression of the training process.
Although this approach demonstrates potential in real-world applications such as drug design \cite{DBLP:journals/ngc/KingSS95}, its performance on the WN18RR link prediction benchmark remains inconsistent. 
This variability primarily stems from the dataset's design, which omits inverse relations in the test set, complicating direct comparisons with baseline methods and suggesting that methods like DFORL could benefit from incorporating negation in rule bodies to enhance the model’s ability to handle higher-arity predicates.
NCRL, a complementary approach by \cite{DBLP:conf/iclr/ChengAS23}, focuses on rule compositionality, sampling alternative paths between connected nodes to construct Horn clauses.
This method employs an iterative algorithm combining an RNN-based selection process with an attention mechanism, aiming to maximize the likelihood that head predicates can be reconstructed from sampled path predicates. 
By leveraging predicate embeddings, NCRL outperforms \cite{DBLP:journals/ai/GaoICW24} on WN18RR, mainly due to its higher ability to infer semantic relationships when inverse relations are missing. 
Its strong performance in low-data regimes and rapid convergence on GPU makes NCRL a compelling alternative for the same task.\\
\indent
Applied to tabular data, Inductive Logic Programming foundations are leveraged in Fold-SE~\cite{DBLP:conf/padl/WangG24} to extract interpretable logical rules for supervised tabular classification. 
Designed to handle both categorical and numerical data with minimal preprocessing, it scales efficiently to large datasets while preserving rule coherence.
Following a sequential covering strategy, Fold-SE iteratively refines rules by optimizing a modified Gini Impurity metric to enhance classification performance.
Built upon Fold-SE, NeSyFOLD~\cite{DBLP:conf/aaai/PadalkarW024} adopts ILP on extracted high level features for an image classification task.
NeSyFOLD derives rules from binary activation masks produced by Convolutional Neural Networks (CNNs), assigning semantic labels to kernels post hoc. 
Its reliance on CNN architectures ---extended and outperformed by modern transformer-based models--- constrains its effectiveness, and the exclusion of exception predicates in semantic labeling further restricts its representational capacity.\\
\indent
In the textual domain, Document-level Relation Extraction (DocRE) is a supervised learning task that identifies relations $r(h,t)$ between entities, where $r$ represents the relation, $h$ is the head entity, and $t$ is the tail entity. 
Both NeSy models and \blackBox approaches have been evaluated on DWIE, a widely recognized benchmark for DocRE. 
JMLR, a novel framework introduced in \cite{DBLP:conf/acl/QiDW24}, integrates DocRE with rule extraction through residual connections.
This approach computes rule support as a weighted sum of atom supports, facilitating the implementation of soft-proof mechanisms that combine relations into HCs.
\cite{DBLP:conf/aaai/JainMKS24} adopted an alternative formulation, treating DocRE as a knowledge base link prediction task and utilizing the \blackBox model DocRE-CLiP to infer relational structures.
FedNSL, introduced in \cite{DBLP:conf/icml/XingL024}, extends NeSy methods to a federated learning setting for DocRE, achieving performance between that of DocRE-CLiP and JMLR. 
While its evaluation of real-world datasets remains limited to DWIE, the framework demonstrates the potential for privacy-preserving algorithms. 
By leveraging variational expectation maximization, it efficiently constrains the rule search space, reducing training rounds while maintaining competitive performance.
A comparative evaluation of these methods ---as shown in Table \ref{tab:benchmarks}--- indicates that JMLR significantly outperforms DocRE-CLiP on DWIE while achieving similar results on other benchmarks, highlighting the advantages of rule-based reasoning in capturing complex multi-sentence relationships.
However, JMLR does not explicitly incorporate external knowledge, a feature more naturally integrated into knowledge-base-driven methods.

\subsection{Natural Logic}
Natural Language Inference (NLI) determines entailment relations between a premise and a hypothesis using formal operators such as equivalence ($\equiv$) and forward entailment ($\sqsubset$).  
\cite{DBLP:journals/tacl/FengYZG22} proposed a transformer-based NeSy model that integrated GPT-2 with reinforcement learning for operator composition and introspective revision using WordNet. 
This approach outperforms baselines like BERT on most benchmarks but remains sensitive to linguistic noise, such as adverbs and prepositional phrase modifications in benchmark datasets.  
QA-NatVer~\cite{DBLP:conf/emnlp/AlyS023} extended this framework to claim verification, constructing proofs through sentence alignment, operator assignment, and a custom DFA. 
As shown in Table \ref{tab:benchmarks}, QA-NatVer highlights the trade-off between NeSy models' explainability and performance, with a notable accuracy drop due to its reliance on a small training set. 
Its \blackBox counterpart, SFAVEL \cite{DBLP:conf/iclr/BazagaLM24a}, overcomes this limitation using self-supervision and distillation to remove the need for labeled data.  
TabVer \cite{aly2024tabver} further extended NLI to tabular claim verification by incorporating numerical reasoning, recognizing equivalences, alongside pragmatic reasoning through deterministic rules. 
While achieving strong results on FEVEROUS datasets, it struggles with exact numerical matching, underscoring ongoing challenges in adapting NeSy methods to less rigid reasoning tasks.

\subsection{Deterministic Finite Automaton}

Deep Q-learning has proven effective in reinforcement learning (RL), particularly in tasks that require sequential decision-making and structured exploration.
Traditional \blackBox methods struggle with environments like the Atari classic game ``Montezuma’s Revenge'', where success depends on complex interactions between objects and precise room navigation.
A Deterministic Finite Automaton provides a structured way to model sequential dependencies by representing state transitions through a finite set of rules. This formalism can be leveraged to guide exploration and improve decision-making in complex RL environments.
Leveraging this formalism, \cite{DBLP:journals/jair/HasanbeigJAMK24} proposes a deep Q-learning framework augmented with DFA synthesis to enhance exploration efficiency and policy optimization. 
By structuring learned behaviors as state transitions, this approach enables more effective reasoning over sequential dependencies.  
Within this framework, the RL algorithm generates exploration traces that capture sequences of state-action pairs alongside reward estimates. 
These traces inform a DFA synthesis module, which formulates logical constraints and employs a SAT solver to construct a minimal-state DFA encoding the agent’s learned behavior.
By leveraging this structured representation, the deep learning module refines policy transitions, optimizing state-action mappings and iteratively improving decision-making within the RL process.
Experiments demonstrate that the framework achieves faster convergence in tasks requiring sequential planning and object interactions, outperforming conventional models that fail to converge.
Integrating expert-designed DFAs further enhances efficiency by eliminating the initial exploration phase and expediting training. 
Future research may extend this approach beyond gaming and investigate using more expressive automata, such as Pushdown Automata, to model hierarchical memory structures \cite{DBLP:books/daglib/0086373}.

\section{Rule-Enforcement Techniques \anchorIconNew[0.15in]} 

This section presents promising methodologies for enforcing constraints in the form of rules over systems for various application purposes. These constraints can be enforced through tailored networks or by regularization.

\subsection{First-Order-Logic}

Integrating First-Order Logic (FOL) rules into constrained image generation has been explored to enhance controllability and trustworthiness.
\cite{DBLP:conf/cvpr/Sueyoshi024} introduced a pioneering approach that extracts FOL constraints from text using dependency parsing and translates them into equations governing the intensity of attention maps. 
Rather than mining logical constraints from large datasets, this method employs a regularization strategy ---a well-established practice in neuro-symbolic reasoning--- as demonstrated by later works in this survey \cite{DBLP:conf/aaai/CaiDCDL22,DBLP:conf/acl/PryorYLKRBG23}.
In this approach, the logical structure is not inferred from data but explicitly imposed through a specialized loss function that guides the generation process. 
Notably, this method natively supports logical quantifiers, which allow for formulating more precise and interpretable constraints. 
Qualitative analyses suggest that incorporating FOL constraints in image generation significantly enhances content fidelity, making this approach highly relevant for applications like chatbots and automated content creation.
However, the absence of a counting mechanism prevents constraint enforcement on repeated entities (e.g., ``a black dog and a white dog''), while a bias toward prototypical examples limits flexibility.
Visual concept learning and semantic parsing could be incorporated to address these challenges \cite{DBLP:conf/iclr/MaoGKTW19}.\\
\indent
Reasoning over images is essential for detecting multimodal misinformation, where deception stems from the interplay of text and visuals.
While traditional \blackBox approaches have achieved high accuracy in this domain, the lack of interpretability remains a major limitation.  
\cite{DBLP:conf/aaai/DongHWJGY024} introduced a NeSy pipeline that enhances transparency by integrating logical reasoning into the classification process.
Using two encoder models, the model extracts patterns from both modalities and employs a teacher-student network to estimate probabilities for three latent variables: image manipulation, cross-modal inconsistency, and image repurposing. 
A unique FOL rule classifies a sample as fake if any of the three variables hold true.
Table \ref{tab:benchmarks} shows this method trailing its \blackBox counterpart by a few points, likely due to the more advanced vision models used by the competitor.\\
\indent
Ensuring neural network reliability is a major challenge, especially in fact-checking~\cite{DBLP:conf/emnlp/BussottiRFMP24}, and media forensics, where errors can have serious consequences.
Formal verification methods guarantee model behavior under specific conditions, ensuring compliance with predefined properties.
\cite{DBLP:conf/ijcai/XieKN22} introduced a structured verification framework based on the Neuro-Symbolic Assertion Language (NeSAL), a fragment of FOL designed for expressing and verifying neural network properties. 
NeSAL formalizes relationships between the inputs and outputs of a neural network under verification (NUV) and those of a specification network, which acts as a binary classifier to assess compliance with a given property. 
For instance, in an image classification task, the specification network outputs ``true'' whenever the NUV correctly identifies a specific class, such as the digit "2" in MNIST.
This framework enables the formal specification and verification of key properties, such as correct classification of specific classes, stability of high-confidence predictions under certain data distributions, and functional equivalence between different architectures within a defined margin of error. 
While NeSAL does not provide quantitative measures for violations, its qualitative evaluations of model confidence under distributional shifts demonstrate its potential for debugging and improving neural networks.
This approach enhances model reliability by providing a structured verification method, making it valuable in high-risk tasks.

\subsection{Probabilistic Logic}
Probabilistic Logic (PL) integrates probability theory with formal logic to model uncertainty, enabling structured reasoning in noisy or incomplete data environments.
SLEER~\cite{DBLP:conf/aaai/CaiDCDL22} addresses reporting bias in temporal event data expressed in text, where uncommon events are overrepresented compared to routine occurrences (e.g., ``It took me an hour to get out of bed'' versus ``It takes me 15 seconds to get out of bed''). 
This imbalance poses challenges for temporal commonsense reasoning tasks, which rely on multiple-choice question formats derived from textual benchmarks. 
SLEER aims to improve model robustness in handling temporal information by incorporating structured reasoning principles.
A language encoder supports multi-task learning, where classification heads predict specific temporal dimensions, each optimized using cross-entropy loss. 
The total loss function integrates these individual losses with a regularization term derived from predefined probabilistic soft logic (PSL) rules. 
Implemented with t-norms, these rules enforce logical consistency by connecting the outputs of the classification heads. Unlike traditional PSL models that use weighted rules, this approach relies on carefully crafted constraints for effectiveness.
Despite its innovative design, the approach yields only marginal improvements on the McTACO \cite{DBLP:conf/emnlp/ZhouKNR19} benchmark\footnote{\url{https://leaderboard.allenai.org/mctaco/submissions/public}}, underperforming on other multi-task encoders such as \cite{DBLP:conf/rep4nlp/PereiraLCAK20}.  
Expanding on this work, \cite{DBLP:conf/aaai/CaiDS00WS23} introduced LECTER, which integrates a context encoder, a logic induction module for temporal dependencies, and a DeepProbLog-based logic validator incorporating human knowledge. 
Optimized through regression and logic entailment losses, LECTER demonstrated strong zero-shot performance on the TIMEDIAL \cite{DBLP:conf/acl/QinGUHCF20} benchmark.
As shown in Table \ref{tab:benchmarks}, LECTER outperforms GPT-3.5~\cite{DBLP:conf/nips/BrownMRSKDNSSAA20} in 2-best accuracy, which checks if the top two ranked answers are correct, while GPT-3.5 is benchmarked with standard accuracy. However, its evaluation remains limited to TIMEDIAL, leaving its generalizability unverified.\\
\indent
Generating interpretable dialogues presents a significant challenge that can be addressed through reasoning over external KBs.
\cite{DBLP:conf/acl/YangZEL22} introduced NS-Dial, a model that integrates external knowledge bases into the dialogue generation process to enhance interpretability.
The framework processes input dialogues through an encoder, which extracts relevant features before a hypothesis generator formulates an initial assumption. 
A hierarchical reasoning engine then identifies supporting triples within the KB, constructing a structured reasoning chain represented as a single weighted HC. 
This inferred knowledge guides the decoder, which generates the next token based on belief scores derived from the reasoning process.
Although this approach improves interpretability, \blackBox methods demonstrate higher performance in dialogue generation---as shown in Table \ref{tab:benchmarks}. 
In the same context, \cite{DBLP:conf/acl/PryorYLKRBG23} proposed a NeSy model to identify dialogue states and transitions in goal-oriented conversations (Dialog Structure Induction), especially in low-resource settings. 
The model combines a Variational Recurrent Neural Network (VRNN) with a symbolic loss based on PSL to enforce structural consistency in sequence prediction. 
Though effective in domain generalization and adaptation, the approach is surpassed by a simpler VRNN baseline as labeled data increases, revealing potential complexity limitations.\\
\indent
PL, like FOL, provides a framework for ensuring network safety. 
\cite{DBLP:conf/ijcai/YangMRR23} proposed a RL policy that enforces action safety through structured probabilistic reasoning. 
The policy uses annotated disjunctions for action probabilities, probabilistic facts for environmental conditions, and logic clauses for safety specifications, enabling queries on shielded policy probabilities and action safety metrics.
The approach employs a PL-based strategy, where a shielded policy gradient optimizes rewards while enforcing safety constraints in situationally aware environments, and a safety gradient enhances action reliability in physically constrained settings like those affected by inertia.
Empirical evaluations indicate that the collaboration between these components is essential to avoid performance degradation.
With RL playing an increasing role in language modeling, we argue that integrating these techniques with instruction tuning may enhance response accuracy and safety in future real-world applications.

\subsection{Deterministic Finite Automaton}

\cite{DBLP:conf/aaai/JiangB0STS21} introduced a temporal-logic-based reward shaping technique to enhance Q-learning in scenarios where the behavior policy deviates from the optimal policy.
Their approach incorporates a potential-based reward term derived from a DFA to encode temporal logic constraints, such as ``following a human'' or ``staying in a corridor''. 
In this framework, the agent iteratively refines a Q-table to estimate state-action values, typically updating it at episode endpoints.
By integrating these rules into the reward function, the method encourages alignment with desired behaviors, accelerating convergence to the optimal policy without enforcing specific actions.
Simulations across various tasks demonstrate that reward shaping is more effective than shielding in exploration-intensive scenarios, while shielding performs better when constraints precisely match task requirements.

\newcommand{\accuracy}{\textsc{Accuracy}}
\newcommand{\fOnemacro}{\textsc{F1-Macro}}
\newcommand{\bleu}{\textsc{Bleu}}

\begin{table*}[ht]
\centering
\small
\begin{adjustbox}{width=\textwidth}
\begin{threeparttable}
\begin{tabular}{l l c c c !{\color{gray!40}\vrule width 1pt} c l}
\toprule
\textbf{Method} & \textbf{Task} & \textbf{Benchmark} & \textbf{Evaluation Metric} & \textbf{Score} & \multicolumn{1}{>{\columncolor{gray!10}}c}{\textbf{Black-Box Competitor}} & \multicolumn{1}{>{\columncolor{gray!10}}c}{\textbf{$\Delta$ Score}} \\
\midrule
JMLR \shortcite{DBLP:conf/acl/QiDW24} & \pickIconNew~Document Lv. Relation Extraction & DWIE \shortcite{DBLP:journals/ipm/ZaporojetsDDD21} & \accuracy & $77.9\%$ & DocRE-CLiP \shortcite{DBLP:conf/aaai/JainMKS24} & $+10.8\%$ \\
QA-NatVer \shortcite{DBLP:conf/emnlp/AlyS023} & \pickIconNew~Claim Verification & FEVER (Dev) \shortcite{DBLP:conf/naacl/ThorneVCM18} & \accuracy & $70.3\%$ & SFAVEL \shortcite{DBLP:conf/iclr/BazagaLM24a} & $-20.0\%$ \\
\hline
NSLM \shortcite{DBLP:conf/aaai/DongHWJGY024} & \anchorIconNew~Image-Text Verification & Weibo \shortcite{DBLP:conf/mm/JinCGZL17} & \fOnemacro & $84.4\%$ & SAFE + Hami-m$^3$d \shortcite{DBLP:conf/mm/Wang0LG024} & $-3.7\%$ \\
SLEER \shortcite{DBLP:conf/aaai/CaiDCDL22} & \anchorIconNew~Temporal Commonsense Reasoning & McTACO \shortcite{DBLP:conf/emnlp/ZhouKNR19} & \fOnemacro & $69.0\%$ & ALICE \shortcite{DBLP:conf/rep4nlp/PereiraLCAK20} & $-10.5\%$ \\
LECTER \shortcite{DBLP:conf/aaai/CaiDS00WS23} & \anchorIconNew~Temporal Commonsense Reasoning & TIMEDIAL \shortcite{DBLP:conf/acl/QinGUHCF20} & \accuracy & $71.5\%$ & GPT-3.5 \shortcite{DBLP:conf/nips/BrownMRSKDNSSAA20} & $+6.5\%^\ast$ \\
NS-Dial \shortcite{DBLP:conf/acl/YangZEL22} & \anchorIconNew~Goal-oriented Dialogue Generation & MultiWoZ 2.1 \shortcite{DBLP:conf/lrec/EricGPSAGKGKH20} & \bleu & $10.6$ & GALAXY \shortcite{DBLP:conf/aaai/HeDZWCLJYHSSL22} & $-9.4$ \\
\hline
NS3D~\shortcite{DBLP:conf/cvpr/HsuM023} & \potionIcon~Visual Question-Answering & Sr3D~\shortcite{DBLP:conf/eccv/AchlioptasAXEG20} & \accuracy & $67.0\%$ & GPS~\shortcite{DBLP:conf/eccv/JiaCYWNLLH24} & $-10.5\%$ \\
\bottomrule
\end{tabular}
\begin{tablenotes}
\item[~]\pickIconNew\hspace{0.8mm}= Rule Mining Task; \anchorIconNew\hspace{0.8mm}= Rule Enforcement Task; \potionIcon\hspace{0.8mm}= Program Synthesis Task.
\item[~]$\ast$\hspace{0.8mm} LECTER \shortcite{DBLP:conf/aaai/CaiDS00WS23} is evaluated using the 2-best accuracy.
\end{tablenotes}
\end{threeparttable}
\end{adjustbox}
\caption{NeSy Competitiveness. The $\Delta$ \textbf{Score} field reports NeSy methods' performance gain over the \blackBox competitors.}
\label{tab:benchmarks}
\end{table*}

\section{Program Synthesis Techniques \potionIcon[0.15in]}
Program Synthesis (PS) techniques generate structured programs in either general-purpose programming languages or domain-specific languages (DSLs) to guide system prediction or behavior. 
This approach shares similarities with methodologies such as NLI syntax, where words and operations are combined to generate a structured representation for each sample.  
However, PS introduces significant complexity, as generating a valid program often depends on semantic parsing or search algorithms like A*. 
Its greater expressive power distinguishes it from formal rule-based methods, which are typically constrained by the removal of quantifiers and face challenges in integrating domains such as mathematics.

\subsection{Context-Free Grammars}
PS has been applied to causal effect estimation, where traditional models often rely on strong assumptions that limit their flexibility. \cite{DBLP:conf/aaai/ReddyB24} introduced NESTER, a model using a DSL defined by a context-free grammar. 
An A* algorithm with heuristic guidance selects grammar rules iteratively, generating differentiable programs acting as inductive biases. 
The DSL supports operations such as binary selection, addition, multiplication, and subset computation but excludes recursion to maintain simplicity. 
While evaluations showed competitive results on small benchmarks, scalability to more diverse datasets remains uncertain.

\subsection{Semantic Parsing}
LINC \cite{DBLP:conf/emnlp/OlaussonGLZSTL23} approaches logical reasoning through semantic parsing, translating natural language inputs into FOL clauses processed by the Prover9 solver.
To enhance robustness, the method employs majority voting across multiple generated programs. 
Comparisons with alternative reasoning paradigms, including naive prediction, scratchpad reasoning, and Chain-Of-Thought, show superior performance on synthetic datasets and competitive results on expert benchmarks. 
A key limitation of this approach is the occurrence of semantic parsing errors, which could be addressed by integrating context-free grammars and consistency-checking mechanisms to improve reliability.
Unlike SATLM \cite{DBLP:conf/nips/YeCDD23}, which encodes problems into SAT formulations and demonstrates strong benchmark performance, LINC relies on FOL-based reasoning, offering a distinct approach to explainability compared to SAT’s declarative framework.\\
\indent
Beyond textual reasoning, NeSy methods have been explored for visual tasks, with NS-CL \cite{DBLP:conf/iclr/MaoGKTW19} serving as a foundational model. 
By integrating a visual encoder, semantic parser, and program executor, it achieves near-perfect generalization on CLEVR \cite{DBLP:conf/cvpr/JohnsonHMFZG17} and strong performance on real-world datasets like VQS \cite{DBLP:conf/iccv/GanLLSG17}, requiring only 10\% of the training data. 
\cite{DBLP:conf/cvpr/StammerSK21} extended this approach by incorporating user feedback and explanation loss, improving robustness against confounding factors through symbolic explanations and Slot Attention mechanisms.
Building on NS-CL, FALCON \cite{DBLP:conf/iclr/MeiMWGT22} introduced meta-learning concepts via hyperboxes, enhancing visual encoders for few-shot learning and biased settings, with competitive results on CLEVR, CUB \cite{wah2011caltech}, and GQA \cite{DBLP:conf/cvpr/HudsonM19}.
NEUROSIM \cite{DBLP:conf/emnlp/SinghGGSGMMKGS23} further explored image manipulation by integrating scene graph-based operations and tailored loss functions, though its real-world applicability remains unclear.
GENOME \cite{DBLP:conf/iclr/ChenSLHG24} shifted the focus to modular incremental learning, generating Python-based components for visual reasoning, image manipulation, and knowledge tagging. 
While outperforming \cite{DBLP:conf/emnlp/SinghGGSGMMKGS23} in real-world tasks, its comparison with \cite{DBLP:conf/iclr/MeiMWGT22} remains limited.
In continual learning, COOL \cite{DBLP:conf/icml/MarconatoBFCPT23} tackled catastrophic forgetting, surpassing regularization-based approaches on CLEVR. 
NS3D~\cite{DBLP:conf/cvpr/HsuM023}, adopting the NS-CL formalism in the 3D setting, demonstrates strong generalization and zero-shot reasoning.
However, as shown in Table \ref{tab:benchmarks}, its performance on the ReferIt3D (SR3D) \cite{DBLP:conf/eccv/AchlioptasAXEG20} benchmark falls short of the \blackBox GPS model \cite{DBLP:conf/eccv/JiaCYWNLLH24}.

\section{Discussion}
Finally, we address open problems, NeSy application challenges, and comparisons with \blackBox methods.

\paragraph{Competing with Black-Box Methods}
NeSy models excel in tasks requiring explicit rule enforcement and structured reasoning, but often struggle in open-domain settings where \blackBox architectures benefit from large-scale, unstructured data.
Table~\ref{tab:benchmarks} highlights these differences, showing that performance inconsistencies across similar benchmarks make some NeSy solutions highly sensitive to their application frameworks.
The Temporal Commonsense Reasoning task exemplifies how score discrepancies may arise from differing benchmarking strategies.
While TIMEDIAL focuses on dialogue masking prediction, McTACO is designed for multiple-choice Question-Answering, resulting in varied performance and further highlighting the sensitivity of NeSy solutions to specific application frameworks.
Performance drops are particularly evident in tasks such as claim verification on FEVER~\cite{DBLP:conf/naacl/ThorneVCM18} and visual question-answering on Sr3D~\cite{DBLP:conf/eccv/AchlioptasAXEG20}, where \blackBox models excel in leveraging semantically-reach unlabeled data.
A key insight from QA-NatVer~\cite{DBLP:conf/emnlp/AlyS023} is the trade-off between interpretability and data efficiency.
Multi-granular chunking and step-by-step scoring improve explainability but limit generalization versus \blackBox methods, creating a balance between symbolic clarity and neural adaptability as NeSy models vary in efficiency and expressiveness.

\paragraph{Future Applications}
As research in NeSy models evolves, certain areas appear to be approaching saturation.
The extensive body of work following NS-CL~\cite{DBLP:conf/iclr/MaoGKTW19} suggests that image-based reasoning has reached a point of diminishing returns.
This evolution demonstrates that well-structured synthetic benchmarks reflecting real-world scenarios can drive field-wide advances (e.g., CLEVR's impact on spatial reasoning).
While reinforcement learning gains traction in robotics and NLP, its high-stakes applications remain limited.
Enhancing trustworthiness through structured constraints with NLI—integrating domain knowledge via DFAs, shaping rewards to enforce desired behaviors, and using shielding to prevent unsafe actions—could enable deployment in critical domains like surgery, medical Q\&A and gene discovery~\cite{DBLP:conf/acl/FrisoniCPMM24,DBLP:journals/cmpb/DomeniconiMMP16}, autonomous driving, and legal analysis.
In this regard, the robust NLI methods proposed in \cite{DBLP:journals/tacl/FengYZG22} would benefit from the rigorous evaluation on benchmarks featuring complex sentence structures with multi-phrase hypotheses, premises, and conclusions, while systematically analyzing alternation, negation, and independence operators—particularly valuable for legal text analysis.
We argue that the design pattern introduced by NeSyFOLD \cite{DBLP:conf/aaai/PadalkarW024}, which treats high-level neural features as semi-structured data for rule mining and partial interpretability, warrants further exploration.
Investigating its application to models like transformers, autoencoders, and GNNs could enable explainability in \blackBox models.
\section*{Acknowledgments}
Research partially supported by: \href{https://aipact-edih.it}{AI-PACT} (CUP B47H22004450008, B47H22004460001); National Plan PNC-I.1 \href{https://www.fondazionedare.it/en/progetto-obiettivi-struttura/}{DARE} (PNC0000002, CUP B53C22006450001); PNRR Extended Partnership \href{https://fondazione-fair.it/en/}{FAIR} (PE00000013, Spoke 8); 2024 Scientific Research and High Technology Program, project ``\href{https://disi-unibo-nlp.github.io/projects/carisbo/}{AI analysis for risk assessment of empty lymph nodes in endometrial cancer surgery}'', the Fondazione Cassa di Risparmio in Bologna; Chips JU \href{https://tristan-project.eu/team/}{TRISTAN} (G.A. 101095947). LG Solution Srl for partially funding a PhD scholarship to G. P. Delvecchio and L. Molfetta.

\section*{Contribution Statement}
All authors contributed equally to this work. For more details, visit \url{https://disi-unibo-nlp.github.io}.

\bibliographystyle{named}
\bibliography{bibliography}

\end{document}